\documentclass[10pt,twocolumn,letterpaper]{article}

\usepackage{cvpr}
\usepackage{times}
\usepackage{epsfig}
\usepackage{graphicx}
\usepackage{amsmath}
\usepackage{amssymb}

\input{def.set}
\usepackage{subfigure}
\usepackage{epstopdf}
\usepackage{booktabs}
\usepackage{natbib}
\usepackage[]{algorithm2e}
\usepackage{multirow}
\usepackage{hhline}
\usepackage{array}
\newcolumntype{x}[1]{>{\centering\arraybackslash}p{#1}}

\usepackage[pagebackref=true,breaklinks=true,letterpaper=true,colorlinks,bookmarks=false]{hyperref}

\cvprfinalcopy 

\ifcvprfinal\fi
\begin{document}

\title{Collaborative Deep Reinforcement Learning for Joint Object Search}

\author{Xiangyu Kong\thanks{Work performed while interning at Microsoft Research.}\\
Peking University\\
{\tt\small kong@pku.edu.cn}
\hspace{-3mm}
\and
Bo Xin\\
Microsoft Research\\
{\tt\small boxin@microsoft.com}
\hspace{-3mm}
\and
Yizhou Wang\\
Peking University\\
{\tt\small yizhou.wang@pku.edu.cn}
\hspace{-3mm}
\and
Gang Hua\\
Microsoft Research\\
{\tt\small ganghua@microsoft.com}
}

\maketitle

\begin{abstract}
We examine the problem of joint top-down active search of multiple objects under interaction, {\em e.g.}, person riding a bicycle, cups held by the table, etc.. Such objects under interaction often can provide contextual cues to each other to facilitate more efficient search. By treating each detector as an agent, we present the first collaborative multi-agent deep reinforcement learning algorithm to learn the optimal policy for joint active object localization, which effectively exploits such beneficial contextual information. We learn inter-agent communication through cross connections with gates between the Q-networks, which is facilitated by a novel multi-agent deep Q-learning algorithm with joint exploitation sampling. We verify our proposed method on multiple object detection benchmarks. Not only does our model help to improve the performance of state-of-the-art active localization models, it also reveals interesting co-detection patterns that are intuitively interpretable.
\end{abstract}

\section{Introduction}
\label{sec:intro}

Given an image, the goal of detecting and localizing objects is to place a bounding box around the instances of a pre-defined object class, such as cars, faces, person/people \cite{felzenszwalb2008discriminatively,viola2004robust,dalal2005histograms,agarwal2004learning}.
With the recent advancement \cite{krizhevsky2012imagenet,simonyan2014very,he2015deep} of deep convolutional neural networks (CNN) on object classification, generic object detection is also attracting more and more attention with fast increasing detection accuracy on popular benchmarks \cite{girshick2014rich,ren2015faster,redmon2015you,liu2016ssd}.

The learning of such deep network based detector is often formulated as the problem of minimizing a loss function over a set of hypothesized target windows, where the loss function contains a classification term and optionally a bounding box regression term.
In the classical sliding window approach \cite{felzenszwalb2008discriminatively,dalal2005histograms,malisiewicz2011ensemble}, the window set often contains hundreds of thousands of windows explicitly sampled on a regular grid at multiple scales. Such methods are prohibitively slow for heavy load state-of-the-art CNN classifiers.

\begin{figure}[t]
    \centering
        {
    \subfigure[Single agent detection]{
        \includegraphics[width=.47\columnwidth]{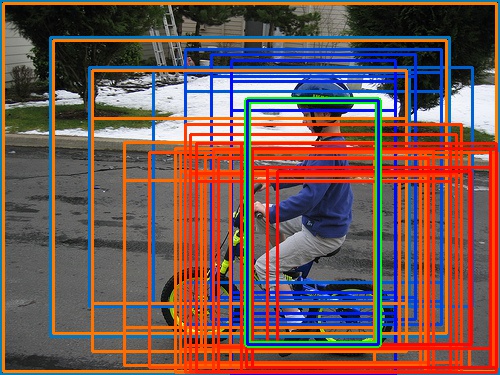}
        }
    \hspace{-1mm}
    \subfigure[Joint agent detection]{
        \includegraphics[width=.47\columnwidth]{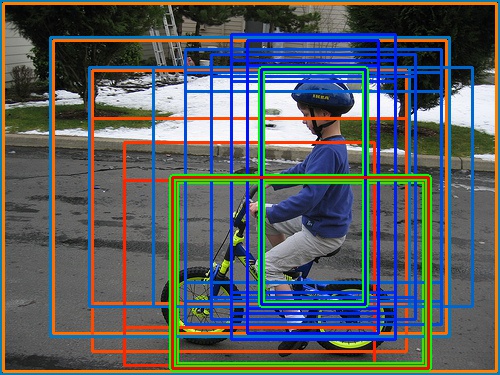}
        }
    }
     \caption{\small{Joint agent detection compared with single agent detection \cite{caicedo2015active}. The bounding box trajectories are indicated by gradual color change. Blue is for person and red is for bicycle. Successful detections are highlighted in bold green. Both objects were detected within 15 iterations by joint detection while single agent detection failed to locate the bicycle even after 200 iterations. (Only the first 30 iterations were illustrated for visualization purpose).
  }}
\label{fig:comp1}
\vspace{-2mm}
\end{figure}

Recent detectors explore the idea of bottom-up object region proposals \cite{girshick2014rich}, where a relatively small set of a few thousand windows were pre-selected \cite{uijlings2013selective} and evaluated. Acceleration were made by sharing computation and pooling over the feature maps from the CNN layers \cite{girshick2015fast,he2014spatial}. These works were further accelerated by integrating the separate region proposal step and the classification step into one network \cite{ren2015faster,liu2016ssd} by using so-called ``anchors" which correspond to regular prototype grid in the image space. However, the number of windows to be evaluated remains several thousand. 
Therefore, the speed of such region-based methods depends on a heavy use of fast GPUs. When computation power is limited, {\em e.g.} only CPUs were available, these pipelines are inevitably slow.

Active search methods provide a promising complementary top-down scheme to reduce the number of windows to be evaluated \cite{mnih2014recurrent,gonzalez2015active,caicedo2015active,yoo2015attentionnet,mathe2016reinforcement}.When searching or localizing objects, biological vision systems are believed to have a sequential process with changing retinal fixations that gradually accumulate evidence of certainty \cite{itti2005neurobiology,larochelle2010learning}. It is therefore highly desirable, both biologically and computationally, to explore computational models that facilitate object search in such top-down behavior. 

Typically, these models learn policies to search for an object by sequentially translating and/or reshaping the bounding box detector. One can view such a search process as an agent searching for the rewarding ground truth bounding boxes and exploit reinforcement learning (RL) algorithms to learn a good policy. In general, these methods can achieve reasonably good performance using only dozens of steps (effectively the number of windows evaluated).

We examine the problem of joint active search of multiple objects under interaction. On one hand, it is interesting to consider such a collaborative detection ``game" played by multiple agents under an RL setting; on the other hand, it seems especially beneficial in the context of visual object localization where different objects often appear with certain correlated patterns, {\em e.g.} person riding a bicycle, cups held on top of the table etc. Such objects under interaction often can provide contextual cues to each other. These cues have good potential to facilitate more efficient search policies.
We make an initial effort to validate such an hypothesis/intuition by devising a computational model.

We present a collaborative multi-agent deep RL algorithm to learn the optimal policy for joint active object localization. Our proposal follows existing wisdom to exploit RL methods but allows for collaborative behaviors among multiple agents in order to utilize contextual information. In this regard, two key questions are open. i) How to make communications effective in between different agents; and ii) how to jointly learn good policies for all agents.

We propose to learn inter-agent communication through gated cross connections between the Q-networks. This is facilitated by a novel multi-agent deep Q-learning algorithm with joint exploitation sampling and a virtual agent based implementation. Finally, we verify our proposed method on multiple object detection benchmarks. Our model helps to improve the performance of state-of-the-art active localization models and it also
reveals interesting co-detection patterns that are intuitively interpretable.

In Section \ref{sec:relate}, we discuss literatures related to our work. In Section \ref{sec:joint}, we
present the details of the proposed cross Q-network structure and a novel multi-agent deep Q-learning algorithm that effectively facilitate training of the crossed Q-networks. In Section \ref{sec:expr}, we present comprehensive experiments on multiple popular benchmarks. Section \ref{sec:concl} concludes this paper. Here, we summarize our major contributions as follows.

\begin{itemize}
\item To our best knowledge, this work presents the first collaborative deep RL solution for joint active object localization.

\item We propose a novel multi-agent Q-learning solution that facilitates learnable inter-agent communication with gated cross connections between the Q-networks.

\item Our proposal effectively exploits beneficial contextual information between related objects and consistently improve the performance of state-of-the-art active localization models.
\end{itemize}

\section{Related Work}
\label{sec:relate}

\noindent
\textbf{Active search.} ~~ The idea of active search for localization is not brand new. To name a few, ``saccade and fixate" biological pattern were explored in the field of visual attention \cite{itti2005neurobiology,larochelle2010learning,wang2011simulating}. In \cite{dollar2010cascaded}, Dollar et al. proposed to estimate pose through cascaded regression steps learnt through gradient descent etc. Latest works on object localization managed to exploit the power of deep learning and achieved more competitive results  \cite{mnih2014recurrent,gonzalez2015active,caicedo2015active,yoo2015attentionnet,mathe2016reinforcement}.

In \cite{mnih2014recurrent}, Mnih et al. proposed a recurrent neural network (RNN) based localization network that accumulatively finds numbers from the cluttered translated MNIST dataset. In \cite{gonzalez2015active}, Garcia et al. proposed to explore statistical relations between consecutive windows and based their model on R-CNN \cite{girshick2014rich} for generic object detection. In \cite{yoo2015attentionnet}, Yoo et al. proposed ``AttentionNet" where at each current window, a CNN was trained to predict quantized weak directions for the next step to simulate a gradual attention shift. In \cite{caicedo2015active,mathe2016reinforcement}, the authors explicitly deployed deep RL and achieved promising performance with much fewer window evaluations than main stream region proposal methods.

However, none of these works examine the problem of joint active search of multiple objects. In order to exploit beneficial contextual information among differnt objects, we present collaborative multi-agent deep RL. We instantiate our idea with Caicedo and Lazebnik \cite{caicedo2015active} as a single active search model baseline, but our mechanism could be applied to other baseline models with minor adaptation.

\noindent
\textbf{Deep reinforcement learning.} ~~ Recently, the field of reinforcement learning revives with the power of deep learning \cite{mnih2013playing,silver2016mastering}. Equipped with effective ideas such as experience replay etc., conventional methods, {\em e.g.} Q-learning, work out very effectively in learning good policies without intermediate supervision for challenging tasks. Our model benefits from these effective ideas in a similar way as recent active methods \cite{caicedo2015active,mathe2016reinforcement} but with specific novel designs motivated by the joint search problem of interest.

Multi-agent machine learning and reinforcement learning are not new topics.
However, conventional collaborative RL methods mostly explore hand-crafted communication protocols \cite{tan1993multi,schwartz2014multi}.
During the preparation of this work, we realize two interesting work that proposed to facilitate learnable communication protocols for multi-agent deep RL \cite{foerster2016learning,sukhbaatar2016learning} and demonstrate superior performance to non-communication counterparts on control management and game related tasks. In \cite{sukhbaatar2016learning}, Sukhbaatar et al. proposed ``CommNet" where policy networks are facilitated with learnable communication channels learnt via back-propagation. In \cite{foerster2016learning}, Foerster et al. proposed ``Differentiable Inter-Agent Learning" to effectively learn communication for deep Q-networks.

Our proposal share the idea of utilizing back-propagation or designing differentiable communication channels but have different cross network structure with gates and a novel joint sampling Q-learning method. Specifically, our cross network structure used explicit gating mechanism to allow a specific agent to be responsible for certain actions. This is motivated by the problem of object search where one agent usually has primary contribution to the policy. Also different from the training of the unfolded RNNs as in \cite{foerster2016learning}, where long range back-propagation may be less effective, our joint sampling design facilitates immediate updates of the parameters and could be easily incorporated into the deep Q-learning algorithm by introducing an auxiliary concept of virtual agent implementation.

\noindent
\textbf{Contextual information for detection.} ~~ Another interesting connection relates to works of pose estimation, landmark detection etc., where one is assigned the task of localizing the positions of all joints or landmarks potentially highly related due to physical constraints. Explicit learning the relationship of different joints/landmarks has been studied in such literatures  \cite{zhu2012face,chen2014articulated}. However similar ideas were rarely explored in general localization problems where such interactions are relatively implicit. Our work partially fills the gap here and proves the concept is similarly applicable in many object-level localization problems.

\begin{figure*}[t]
    \centering
    \includegraphics[width=1.8\columnwidth]{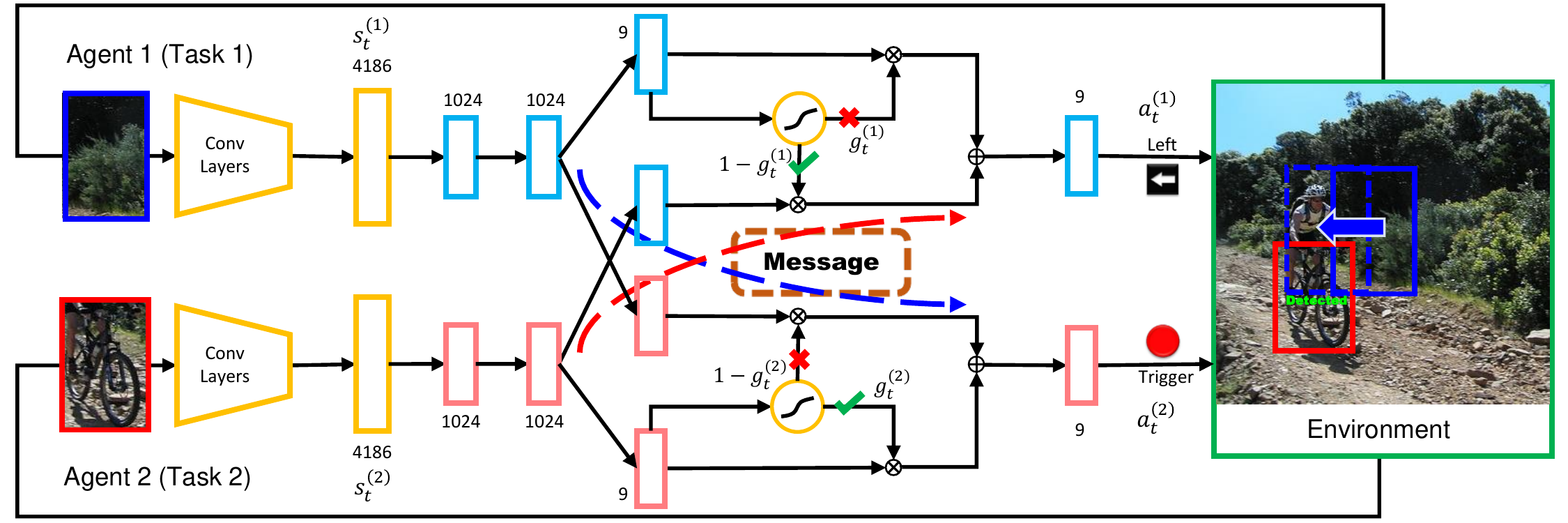}
     \caption{\small{Joint Q-network with gated cross connections and the collaborative reinforcement learning pipeline.}}
\label{fig:joint}
\end{figure*}

\section{Collaborative RL for Joint Object Search}
\label{sec:joint}

We start by recalling a state-of-the-art (single agent) RL method for object localization \cite{caicedo2015active}.

\subsection{Single Agent RL Object Localization}
\label{ssec:single}

Reinforcement learning provides a formal framework concerned with how agents take actions in an environment so as to maximize some notion of cumulative reward.
Formally, RL defines a set of actions $A$ that an agent takes to achieve its goal; a set of states $S$ that represents the agent's understanding/information of the current environment; and a reward function $R$ that helps to learn an optimal policy to guide the agent's actions based on its states.

In \cite{caicedo2015active}, the entire image is viewed as the  environment. The agent transforms a bounding box according to a set of actions. The goal of the agent is to land a bounding box at the target object's location.
Specifically, the set of actions were defined as follows. $\calA:=$ $\{$\textit{move right, move left, move up, move down, scale bigger, scale smaller, aspect ratio change fatter, aspect ratio change taller, trigger} $\}$. Each action makes a discrete change to the box by a factor relative to its current size. The action \textit{trigger} means that the agent thinks it finds the object.

The state representation is defined as a tuple $s:=(o,h)$. $o$ is a feature vector of the observed region (plus some extra margin for context) extracted from a CNN layer, and $h$ is a fixed-size vector of the action history.
The concatenation of $o$ and $h$ is fed into a typical Q-network of two fully connected layers. The network outputs a 9-dimensional vector corresponds to nine action choices. In Figure \ref{fig:joint}, the networks shown in the same color {\em e.g.} in blue/red provide illustrations of this architecture.

The reward function $R(a,s\rightarrow s')$ is defined for an agent when it takes the action $a$ to move from state $s$ to $s'$.
\begin{equation}
    R(a,s\rightarrow s') = sign(IoU(b',g)-IoU(b,g))
\end{equation}
where $IoU(b,g)=area(b \cap g)/area(b \cup g)$ is the  Intersection-over-Union (IoU) between the target object bounding box $g$ and the predicted box $b$.

With the action set, state set and reward function defined, the authors in \cite{caicedo2015active} directly applied deep Q-learning \cite{mnih2013playing} to learn the optimal policy.
More details on setting parameters can be found in \cite{caicedo2015active}.
They also proposed an interesting design for setting masks in the image after taking the trigger action. This design allows for effective detection of multiple instances of the same class. Finally, the authors applied a post SVM classifier to all windows in the trajectory to boost performance.

\subsection{Collaborative RL for Joint Object Localization}
\label{ssec:joint}

We generalize the single agent RL model for joint object search. The key concepts include gated cross connections between different Q-networks; joint exploitation sampling for generating corresponding training data, and a virtual agent implementation that facilitates easy adaptation to existing deep Q-learning algorithm.

\subsubsection{Q-Networks with Gated Cross Connections}
\label{sssec:cross}

Specifically, Q-learning is an RL algorithm used to find an optimal action-selection policy.
The Q-function (action-value function) of a policy $\pi$ is defined as $Q^{\pi}(s,a)=\mathbb{E}[R_t|s_t=s, a_t=a]$ where the subscribes of $t$ denote the time step. The optimal action-value function obeys the Bellman optimality equation $Q^*(s,a)=\mathbb{E}_{s'}[r+\gamma \max_{a'} Q^*(s',a')|s,a]$ where $r=R(a,s \rightarrow s')$ is the specific reward by taking action $a$ to move state $s$ to $s'$ and $\gamma\in [0,1]$ is a discount factor for future returns.

Deep Q-learning \cite{mnih2013playing} uses deep neural networks to represent the Q-function, {\em i.e.} $Q(s,a;\theta)$ where $\theta$ is the network parameters. (A common choice of the Q-network consists of two fully connected layers as illustrated in Figure \ref{fig:joint}.)
Note that, suppose for each agent $i$ we instantiate one Q-network $Q^{(i)}(a^{(i)},s^{(i)};\theta^{(i)})$, in the setting of multi-agent RL, one would naturally desire a Q-function (with a slight abuse of notation, we keep using Q-function here) that facilitates inter-agent communication $Q^{(i)}(a^{(i)}, m^{(i)},s^{(i)}, m^{(-i)};\theta^{(i)})$ where $m^{(i)}$ denotes some form of messages sent out from agent $i$ and $m^{(-i)}$ denotes messages received from other agents.

Conventionally, $m$ is often hand crafted based on prior knowledge about the actions and the states. This can be formalized as a function of $m(a,s;\theta_m)$ where $\theta_m$ is manually designed. Therefore, a natural idea would be to construct differential messages where $\theta_m$ could be learned via gradient back-propagation.
This idea is intuitive and reasonable in the same sense of many deep learning successes where learnable features outperform hand crafted ones.

Specifically, we define an agent-wise Q-function as
\begin{equation}
\label{eq:q1}
   Q:=Q^{(i)}(a^{(i)},m^{(i)},s^{(i)},m^{(-i)};\theta_a^{(i)},
\theta_m^{(i)}),
\end{equation}
where $\theta_a$ and $\theta_m$ represents parameters related to actions and messages respectively.

We would now argue that when Q-function were parameterized with deep networks, there are intuitively to the order of $L^2$ ($L$ is the number of layers of the Q-network) possible configurations for us to construct message channels. This is because the messages could be emitted and received at any layers. Moveover, there should be no global optimal configuration, instead suitable configuration of the message channel should be selected in a problem-dependent manner.

We notice two recent work also propose to facilitate learnable communication protocols for multi-agent deep RL \cite{foerster2016learning,sukhbaatar2016learning} applied to control management and game related tasks respectively. However, we notice that one important insight is missing from the current trend. Messages are often taken-in
in a non-discriminative manner and merged with the information flows in the network directly. Actually, allowing the messages to go through an explicit learnable gate (as did in LSTM cells \cite{hochreiter1997long}) helps better merging  the information and facilitates agent-responsible actions.

The idea is motivated from the object search problem of our interest. In general, when searching for a specific object, we would like the agent in charge of detecting the target class to be a primary source of making decisions. Meanwhile we also want to allow other agents to contribute their advices especially when the primary source feels confused in certain situations. Learnable gating mechanism is a natural fit.

Specifically, we design our cross Q-network message channels as illustrated in Figure \ref{fig:joint}.  We add cross connection from the penultimate layer between Q-networks of different agents. We denote the output from this layer of the Q-network of agent $i$ as $\bx_{L-1}^{(i)}$. We then have
\begin{equation}
\label{eq:gate}
\begin{split}
  \bar{\bx}^{(i)} & = \sigma(\bW^{(ii)} \bx_{L-1}^{(i)}+\bb^{(ii)}) \\
  g^{(i)} & = \sigma(\bW^{(ig)} \bar{\bx}^{(i)} + \bb^{(ig)}) \\
  \bm^{(i)} & = \sigma(\bW^{(im)} \bx_{L-1}^{(i)}+\bb^{(im)})
\end{split}
\end{equation}
where $\sigma$ represent the sigmoid function such that $\sigma(z)=1/(1+\exp(-z))$.

Now instead of directly inputting $\bx_{L-1}^{(i)}$ to the next layer as in the single agent case, we also take in the messages from other sources weighted by gates and define
\begin{equation}
  \bx_{L}^{(i)} = g^{(i)} \cdot \bar{\bx}^{(i)} + (1-g^{(i)}) \cdot \bm^{(-i)}
\end{equation}
Note that, the sigmoid function tends to push the output to approximately $0$ or $1$. Therefore, with this simple gating induced, we are able to learn effective agent-responsible decisions.
This helps us to better understand the searching process. Moreover, now that many actions were effectively determined by one primary agent (and so will the corresponding gradient updates discussed later), one can directly apply learnt networks even when other agents do not co-exist.

\subsubsection{Joint Exploitation Sampling}
\label{sssec:jsample}

We now turn to the problem of jointly training all Q-networks. Since we do not have any immediate supervision in an RL setting, we cannot directly back propagate gradients in a multi-task manner. The key idea is to jointly sample the next steps during the exploitation phase.

Specifically, in the case of a single agent, in order to reach the Bellman optimality, the Q-learning algorithm proceeds in an iterative fashion. At each iteration, one would sample/choose an action $a_t$ according to the current estimate of the Q-function. One then executes this action $a_t$ in the emulator and observes reward $r_t$ and state $s_{t+1}$. After this, one updates the parameters of the Q-function by minimizing the distance of $(Q(a_t, s_t;\theta)-(r_t+\gamma \max_{a'} Q(a',s_{t+1};\theta^-)))^2$. Here $\theta^-$ are the parameters of a target network. $\theta^-$ can be a copy of the online network but often is another network frozen  for a number of iterations while one updates the online network $Q(a,s;\theta)$ \cite{mnih2013playing}.

In the multi-agent setting, we propose to sample the action $a_t^{(i)}$ of agent $i$ according to both the activations of itself and the messages from other agents. We jointly perform such sampling to all the agents. For instance, in Figure \ref{fig:joint}, this corresponds to a joint feed-forward pass from both networks. These samples are later used to update all parameters by jointly minimizing the following distance for all $i$.
\begin{equation}
\begin{split}
 L^{(i)}:=( Q^{(i)}(a_t^{(i)}, m_t^{(i)}, s_t^{(i)},m_t^{(-i)};\theta_a^{(i)}, \theta_m^{(i)})-\\(r_t^{(i)}+\gamma \max_{a'^{(i)}} Q(a'^{(i)}, s_{t+1}^{(i)};\theta_a^{(i)-},\theta_m^{(i)-})))^2
\end{split}
\end{equation}
Since the messages are also differential, joint minimization of the above functions will update parameters related to each of the agents as well as all the message channels in-between. Specifically, the gradient updates of $\theta_a^{(i)}$ comes from the loss of itself {\em i.e.} $L^{(i)}$, while the gradient updates of $\theta_m^{(i)}$ comes from the loss of other agents {\em i.e.} $L^{(-i)}$.

Note that, in principle we could view all agents under one global Markov decision process (MDP) assumption and search for an optimality in the joint action space using the regular Q-learning algorithm. The flip side of this choice, however, is a much larger searching space ($81$ v.s. $18$ in the two agent case) that may require combinatorially much more training data and time. In this regard, the proposed joint sampling strategy can be viewed as an upper-bounding approximation to global optimal. However, we observe that this proposal effectively facilitates gradient back-propagation to all the parameters and can jointly learn good policies for all the Q-networks as desired.

\subsubsection{Virtual-Agent Implementation of Joint Training}
\label{sssec:virtual}

\begin{figure}[h]
    \centering
    \includegraphics[width=0.9\columnwidth]{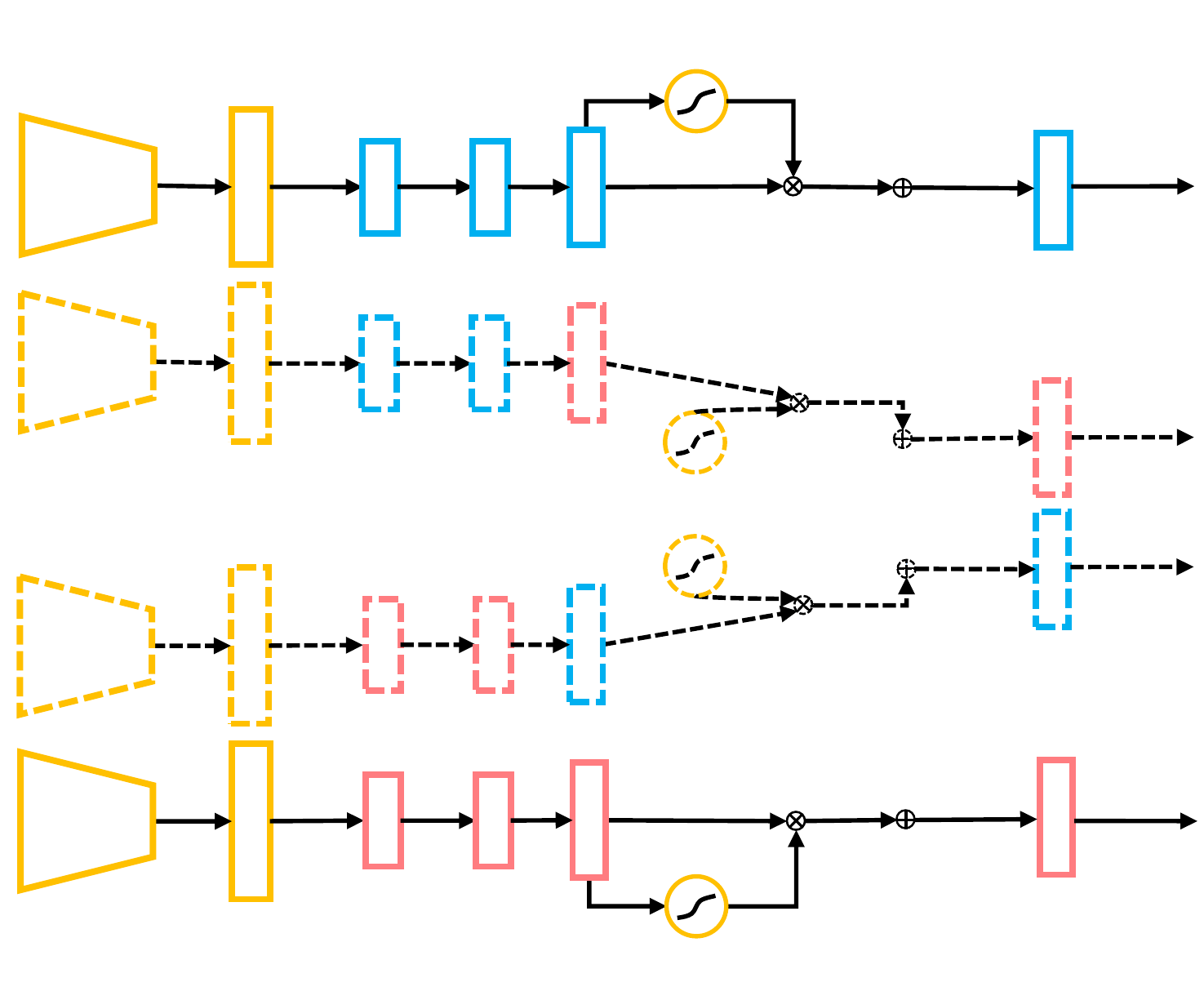}
     \caption{\small{An illustration of the actual and virtual agents of the example in Figure \ref{fig:joint}. Each row represents one agent and the dashed ones in the middle are virtual agents.}}
\label{fig:virtual}
\end{figure}

Intuitively the joint sampling idea can be implemented via simultaneously forward and backward passes through all Q-networks. However in practice, we adopted an alternative implementation with a concept of virtual agents. For each Q-network of an object class, we assign an actual agent detector. Meantime, for each cross network connection we assign a what we call virtual agent. The virtual agents share weights of the corresponding layers with the actual agents. Figure \ref{fig:virtual} illustrate this idea for the example of Figure \ref{fig:joint}.

There are two major advantages of this implementation. 1) By considering agents in such a separate manner (and share weights afterwards), we can easily incorporate our design to almost all existing RL algorithms. One can simply implement an extra outer for-loop for all agents followed by necessary weight copying steps. 2) More importantly, this also allows each agent, including virtual ones, to maintain its own pool (replay memory \cite{mnih2013playing}) of samples. These samples are used for updating the corresponding parameters. Note that in modern RL algorithms with deep networks, the concept of replay memory pool are extremely important for stabilizing the learning process.

For example, suppose we would like to jointly train person and bicycle detectors. During training, we have images that contain both classes $D_{both}$ and also images that only contain either person $D_{person}$ or bicycles $D_{bicycle}$. Benefit from an agent-wise replay memory as proposed, the actual person and bicycle agents could be effectively trained with data from $D_{both} \cup D_{person}$ and $D_{both} \cup D_{bicycle}$ respectively, while the cross connections (represented by virtual agents) are only trained with data from $D_{both}$ as desired.

Finally, we update the denotation of the Q-functions in the context of the virtual agent implementation as follows.
\begin{equation}
\begin{split}
   &Q_{a}^{(i)}(a^{(i)},s^{(i)};\theta_{share}^{(i)}, \theta_{self}^{(i)}); \\
   &Q_{v}^{(i\rightarrow j)}(a^{(i)},s^{(i)};\theta_{share}^{(i)},
\theta_{self}^{(i\rightarrow j)}).
\end{split}
\end{equation}
The main changes from the definition in Equation (\ref{eq:q1}) are to use $\theta_{self}^{(i\rightarrow j)}$ to replace the conceptual out-message $\bm^{(-i)}$ and to use post addition to replace the conceptual in-message $\bm^{(i)}$. (Note that, as illustrated in Figure \ref{fig:virtual}, we put the gating part inside the Q-function by definition.) Specifically, we summarize the final multi-agent Q-learning algorithm with joint sampling and virtual agent in Algorithm \ref{alg:alg}.
Although the algorithm applies in general cases, we usually consider only two object classes at the same time, therefore the number of virtual agents is very controllable.

\setlength{\textfloatsep}{10pt}
\begin{algorithm}[t]
\caption{Multi-agent Q-Learning Algorithm}
\label{alg:alg}
{
	Initialize replay memory of all agents $D^{(i)}$\;
	Initialize all Q-networks with random weights (or potentially with pre-trained networks)\;

\For{episode = 1,M}{
Initialize sequence $s_1^{(i)}=\phi({x_1})$ for all $i$\;
		
\For{t=1,T}{
With probability $\epsilon$ select a random action $a_t^{(i)}$,
otherwise select
$
\begin{array}{cc}
  a_t^{(i)} = \argmax_a \{ Q_{a}(a,s_t^{(i)};\theta_{share}^{(i)}, \theta_{self}^{(i)})
  \cr  +  \sum_{j\neq i}{Q_{v}(a,s_t^{(j)};\theta_{share}^{(j)}, \theta_{share}^{(j\rightarrow i)})} \}
\end{array}
$\;
Execute action $a_t^{(i)}$ in emulator and observe reward $r_t^{(i)}$\;
Set $s_{t+1}^{(i)}$ with $s_t^{(i)},a_t^{(i)}$\;
Store transition $\left(s_t^{(i)}, a_t^{(i)},r_t^{(i)},s_{t+1}^{(i)}\right)$ in $D^{(i)}$ and $D^{(j\rightarrow i)}$ for all $j$\;
Sample random mini-batch of transitions $\left(s_{t'}^{(i)}, a_{t'}^{(i)},r_{t'}^{(i)},s_{t'+1}^{(i)}\right)$ from $D^{(i)}$\;
Set $y_{t'}^{(i)}=$
$
\begin{cases}  r_{t'}^{(i)} ~~~~~if~terminates ~at ~t'+1 \\  r_{t'}^{(i)} + \gamma \max_{a'}\hat{Q}_a(a_{t'}^{i}, s_{t'+1}^{(i)};\theta^{(i)-}) ~~else \end{cases}
$\;
Perform a gradient descent step on $(y_{t'}^{(i)} - Q_a(a_{t'}^{(i)}, s_{t'+1}^{(i)};\theta_{share}^{(i)}, \theta_{self}^{(i)}))^2$ with respect to $\theta_{share}^{(i)}, \theta_{self}^{(i)}$\;
Copy $\theta_{share}^{i}$ to all virtual agents $(i\rightarrow j)$\;
\For{$j\neq i$}{
Sample mini-batch from $D^{(j\rightarrow i)}$\;
Update $\theta_{share}^{(j)}, \theta_{self}^{(j\rightarrow i)}$ of the virtual agents $Q_v^{(j\rightarrow i)}$ as above\;
Copy  $\theta_{share}^{j}$ to actual agent $j$\;
}
}
}
}
\end{algorithm}

\section{Experiments}
\label{sec:expr}

\subsection{Data Construction}

Although different classes of objects co-exist in many situations in real life, there are few datasets explicitly collect data for joint detection tasks.
However, we notice that many images from popular detection datasets such as the PASCAL VOC datasets and the COCO dataset have labeled objects of different classes and these images were categorized under all related classes.
These images naturally provide a source for us to construct some useful datasets  to validate our hypothesis and methods.
Specifically, we selected: \{\textit{person+bicycle (VOC), ball+racket (COCO), person+handbag (COCO), keyboard+laptop (COCO)}\}. With these pairs, we construct two datasets for evaluation purpose.
$D_1$ consists of images that only contain one object for each class. This dataset is used to prove certain concepts since learning and testing tend to be more effective on this relatively cleaned dataset. $D_2$ consists of all images of the person and bicycle categories from the PASCAL VOC datasets. This one is used to evaluated our proposed method against results of existing single agent models.

\begin{figure*}[t]
    \centering
    {
    \includegraphics[width=1.9\columnwidth]{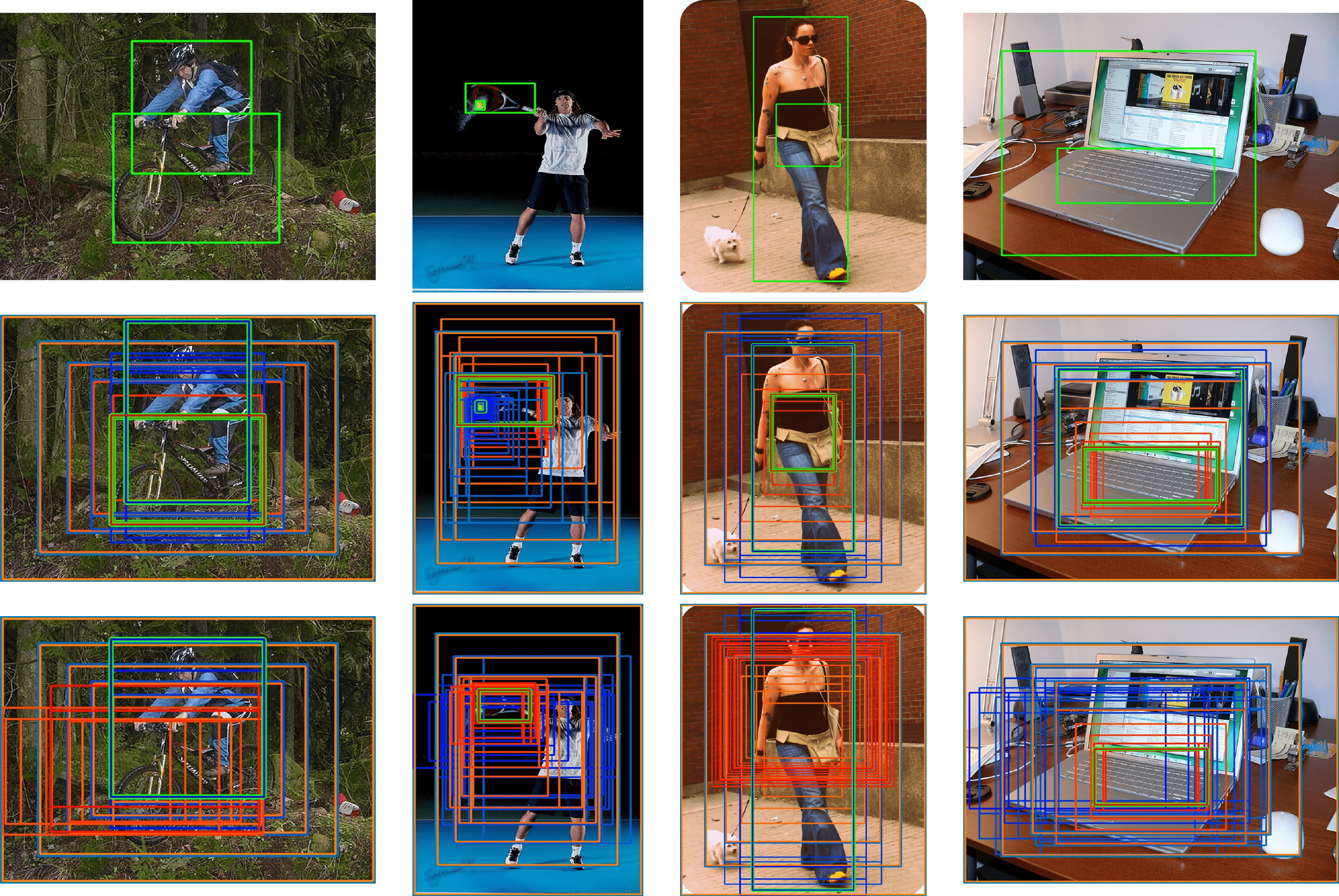}
    }
     \caption{\small{Joint agent detection (mid) compared with single agent detection (bottom). The bounding box trajectories are indicated by gradual color change with blue and red each for one detector. Successful detections are highlighted in bold green.}}
\label{fig:success}
\end{figure*}

\subsection{Details of implementation}
\label{ssec:detail}

For comparison purpose, we implemented the single agent model precisely according to \cite{caicedo2015active}. Specifically, we use a CNN architecture of the ZF-net \cite{zeiler2014visualizing} as the CNN feature extractor and use the layer fc6 (of dimension 4096) as the feature vector $o$. (More advanced CNN models such as \cite{simonyan2014very,he2015deep} could also be used to potentially further increase detection accuracy, but they are not the focus of our study in this paper.) Action history $h$ encodes 10 past actions and therefore has a dimension of 90. $s$ is then of dimension 4186.
The Q-network consists 2 fully connected layers of dimension 1024 and the final output is a 9 dimensional vector representing Q-values for 9 actions. $\epsilon$-greedy training were applied to balance between exploration and exploitation.
During testing, a maximum of 200 steps are used. For more details, please refer to \cite{caicedo2015active}. Since the authors of \cite{caicedo2015active} did not release their code, we implemented our own version. We manage to have achieved very close performance as reported in \cite{caicedo2015active} though not exactly the same. The differences may be due to the randomness involved in sampling.

In the case of multiple agents, cross connections between Q-networks are implemented as a fully connected layer from one agent's penultimate layer to another agent's last layer with a post multiplication by a scalar gate as defined in Equation (\ref{eq:gate}). The dimensions are consistent with the corresponding layers in the single agents. For joint training, we initialize each actual single agent network using pre-trained models and initialize cross connection with random weights.
We applied the $\epsilon$-greedy strategy of \cite{caicedo2015active} where we have tuned the learning rates to achieve better convergence in our case.
We report detection results from the joint model on dataset $D_1$ since it contains both classes by construction; and report detection results using fine-tuned single agent model by joint training, which demonstrates the ability of the gating mechanism to facilitate agent specific inference and learning.

\subsection{Improvement over Single Agent Methods}

In Table \ref{tab:test1} and Table \ref{tab:test2}, we demonstrate the performance of our proposal when compared with single agent models.
Our joint model consistently outperforms the single agent model on dataset $D_1$. We notice that on the combinations of \textit{person+bicycle (VOC)} and  \textit{laptop+keyboard (COCO)}, the improvement is much more obvious. This is because the configuration of these combinations are relatively more stable across images, {\em e.g.} person riding a bike and laptop contains the keyboard etc. Meanwhile, the configurations of \textit{person+handbag (COCO)} and  \textit{ball+racket (COCO)} have multiple modes in all the images and more ``noisy" images that contain little information for co-localization.

\begin{table}[h] \small
\caption{Localization accuracy on $D_1$. Top: single, bottom: joint.}
\label{tab:test1}
\begin{center}
{
\begin{tabular}{c|c|c|c|c|c|c|c}
\hline
\multicolumn{2}{c|}{(VOC)} & \multicolumn{2}{c|}{(COCO)} & \multicolumn{2}{c|}{(COCO)} &  \multicolumn{2}{c}{(COCO)} \\
\hhline{--------}
 \!\!person\!\! & \!\!\!\!bike\!\!\!\! & \!\!\!\!ball\!\!\!\! & \!\!racket\!\!  &   \!\!person\!\! &  \!\!hbag\!\!  & \!\!laptop\!\! &  \!\!kboard\!\!  \\
\hline
  76.9 &  61.5 & 52.0 & 59.3 & 80.4  & 45.1  & 60.6   & 56.9 \\
 \textbf{86.0} & \textbf{77.8}  & \textbf{53.9} & \textbf{60.2} & \textbf{82.5} &  \textbf{46.4}  & \textbf{64.6} &  \textbf{64.7}  \\
\hline
\end{tabular}
}
\end{center}
\vspace{-3mm}
\end{table}

\begin{table}[h] \small
\caption{Localization accuracy on $D_2$.}
\label{tab:test2}
\begin{center}
{
\begin{tabular}{c|c|c}
\hline
$D_2$ & person (VOC)  &  bicycle (VOC) \\
\hline
Mathe et al. \cite{mathe2016reinforcement} &   18.7  & 31.4   \\
Caicedo et al.  \cite{caicedo2015active} &  45.7 &  61.9     \\
Ours (Single)  &   44.6   &   62.2   \\
Ours (Joint) &   \textbf{45.6}  &   \textbf{63.9} \\
\hline
R-CNN \cite{girshick2014rich} &  54.2  &  69.7 \\
\hline
\end{tabular}
}
\end{center}
\vspace{-2mm}
\end{table}

When tested on dataset $D_2$, our joint model also achieved better performance than single active search models. The performance gain is moderate in this case. This is because the number of images containing both object classes is small when compared with that for each category, therefore the extra information gain is diluted. This is especially the case for the person category whose number of images is much larger.

Note that, as a comparison, state-of-the-art detection models such as R-CNN \cite{girshick2014rich} and its extensions can achieve better results using bottom-up region proposals with much more windows. In Section \ref{ssec:recall}, we will demonstrate that the recall of such region proposal methods will drop significantly when only limited number of proposals (namely, the actually computation) are allowed. Therefore, an effective combination of both the top-down active search methods and the bottom-up region proposal methods will help in striking a balance between good performance and limited computation. We leave this for our future work.

In Figure \ref{fig:success}, we illustrate the search process in some example images. In these cases, while our joint detection model can successfully locate objects from both categories, the single agent model often can only detect one or neither of them correctly.
The locations of the final bounding boxes from the joint model also seem better overlapped with the ground truth objects.
Moreover, the number of steps taken by the joint model is much smaller. For example, from top to bottom, the illustrated number of steps for our model are: 10, 24, 7 and 11 respectively. We show the first 30 steps of the single model for visualization purpose. Actually, in all these three cases, the single agent model failed to locate both objects within 200 steps.
In practice, our model only uses several tens of steps to locate both objects and the number of steps are often less than when using two single agents, which has already been shown to be consistently superior to region proposal methods when using limited number of proposals \cite{caicedo2015active}.
We provide more quantitative analysis in Section \ref{ssec:recall}.

\setlength{\textfloatsep}{20pt}
\begin{figure}[t]
    \centering
    {
    \includegraphics[width=1\columnwidth]{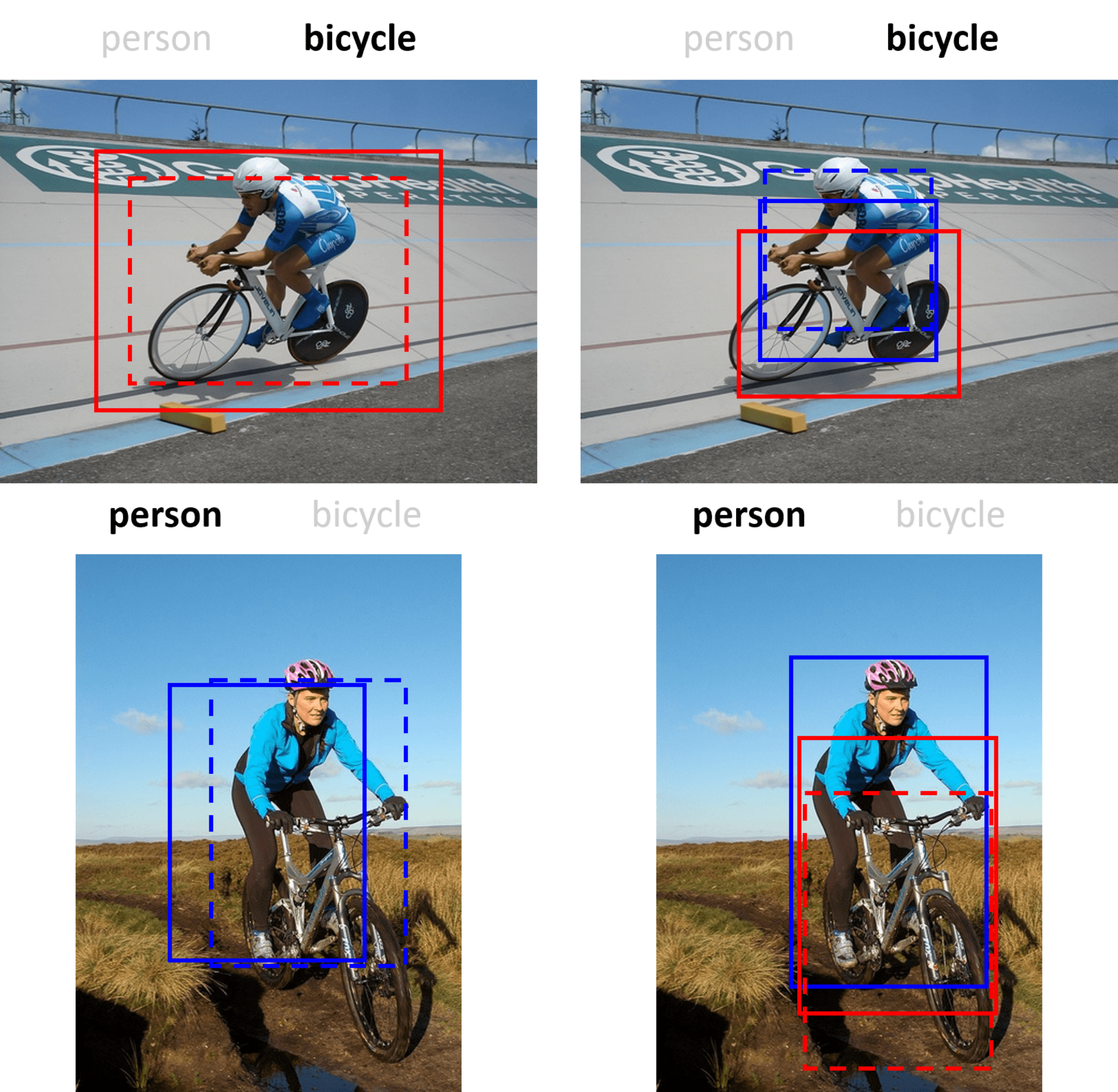}
    }
     \caption{\small{Examples of actions dominated by specific agents. The solid bounding boxes illustrate the current positions of each detector. The dashed bounding boxes illustrate the next positions and indicate the corresponding actions. Blue is for person and red is for bicycle. The agent which dominates the choice of action (by checking the gate value) are highlighted in bold black.}}
\label{fig:step}
\end{figure}

The agents in a joint model help each other in a rational fashion. For example, in the first column of Figure \ref{fig:success}, the bicycle looks relatively less distinguishable from the background of bushes. While the single bicycle agent fails to locate its target, in the joint model, the detection of the person seems to help locate the bicycle since it often presents the pattern of a person riding a bicycle.
In the second column, the tennis ball looks very small and a single tennis ball agent has trouble finding it; meanwhile benefit from the co-existing pattern with the racket learnt by the joint model, we can successfully detect the ball. The third and fourth column also demonstrate cases where a relatively easy-to-detect object (person and keyboard in these cases) helps to locate the more challenging ones (bag and laptop) due to learnt co-existing patterns.

\subsection{Analysis}
\label{sec:analysis}

\subsubsection{Step by Step Examination}

In Figure \ref{fig:step}, we demonstrate some examples of actions, the choice of which were dominated by specific agents. As the left two images show, when the clue of the primary agent is clear, the actions are often taken according to themselves. For example, given their current input bounding boxes, the bicycle agent knows to scale smaller in the top left image and the person agent knows to move right in the bottom left image.

However, in cases where the primary agent is less confident of itself, our proposal effectively queries information from other agents. For instance, in the bottom right image, the bicycle detector were pushed down, but this action is primarily decided by the person agent. This is probably because the person detector has triggered a target and it feels more certain about the situation. Due to the learnt pattern, it fires relatively strong signals indicating a bicycle underneath and helps to push the red box downwards. Of course this does not necessarily mean the primary bicycle agent has to make the wrong choice of actions, but simply it may be less confident given its relatively noisy current input. 

\subsubsection{Evaluation of Recall}
\label{ssec:recall}

Note that, for active search methods, all the regions attended by the agents can be understood as object proposal candidates. In all our experiments, we have used a maximum of 200 steps following \cite{caicedo2015active}, however, this is oftentimes not necessary. In practice, our model usually only uses several tens of steps to locate both objects and the number of steps are often less than when using two single agents.

\begin{figure}[h]
	\centering
	{
		\includegraphics[width=0.75\columnwidth]{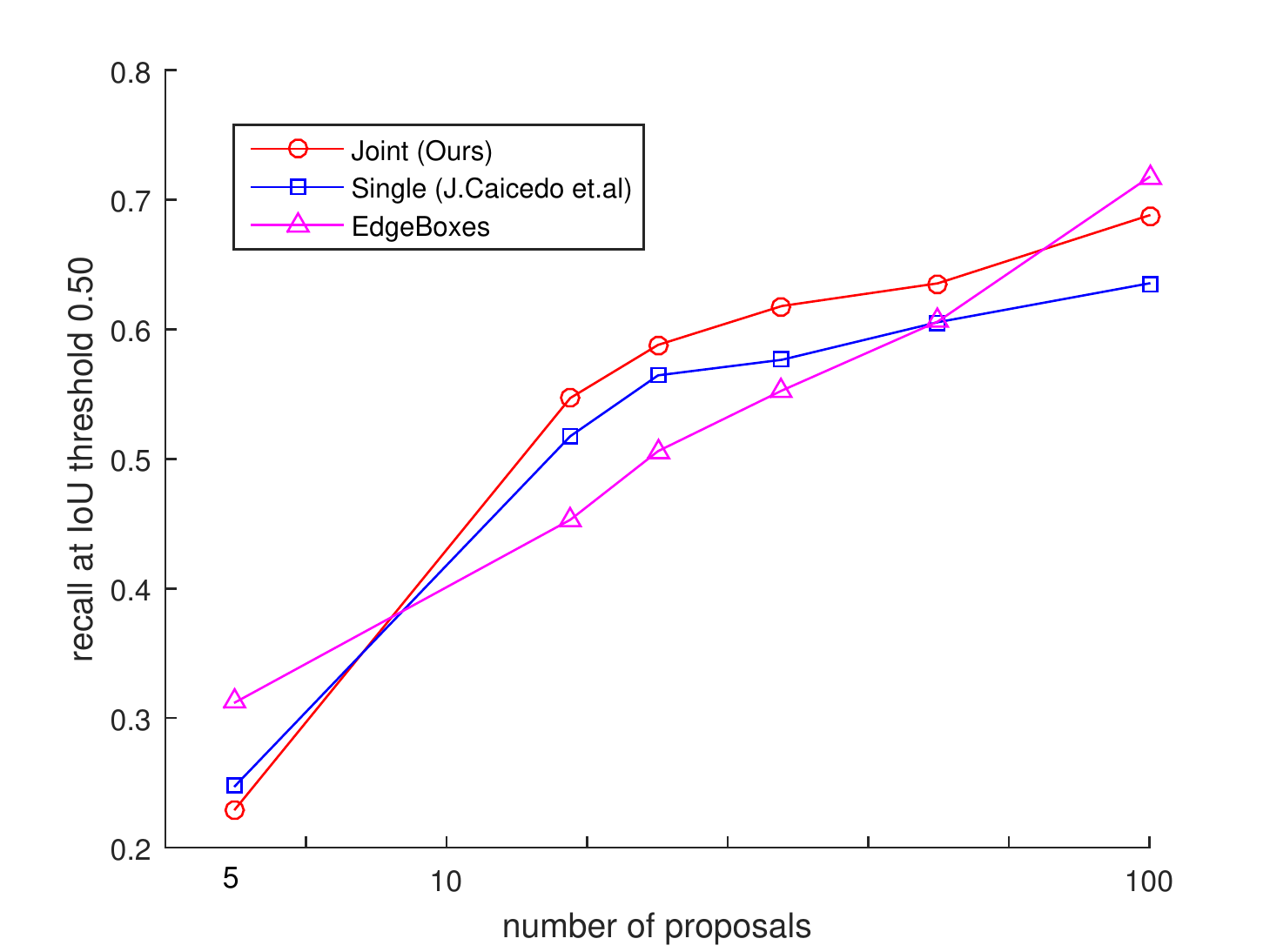}
	}
	\caption{\small{Recall as a function of the number of proposed regions. Compared with region proposal methods, active search methods are better at early recall: only several tens of proposals per image reach 50\% recall. Our joint model is even better than the single agent model. }}
	\label{fig:recall1}
\end{figure}

\begin{figure}[h]
	\centering
	{
		\subfigure[bicycle]{
			\includegraphics[width=.85\columnwidth]{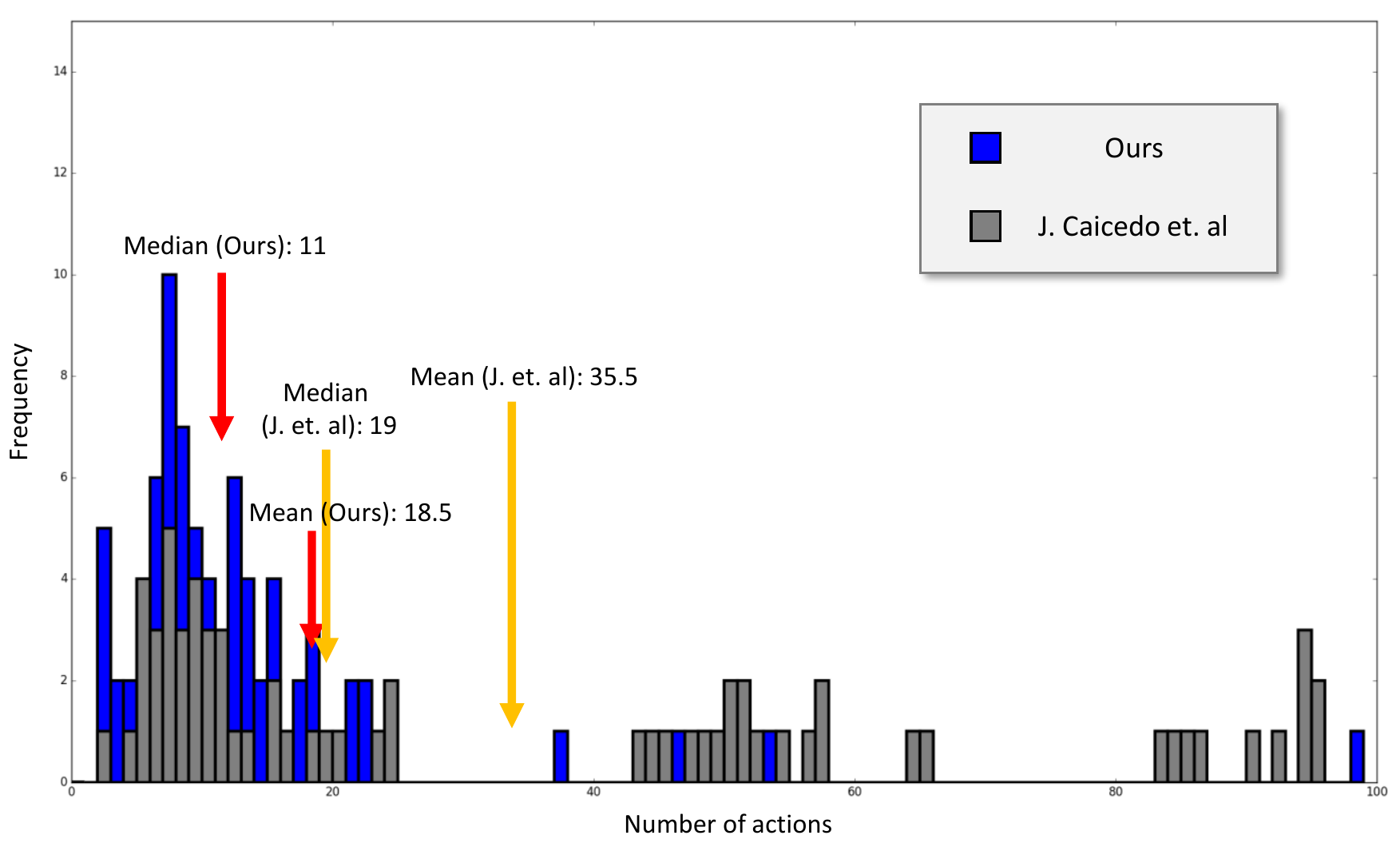}
		}
		\subfigure[person]{
			\includegraphics[width=.85\columnwidth]{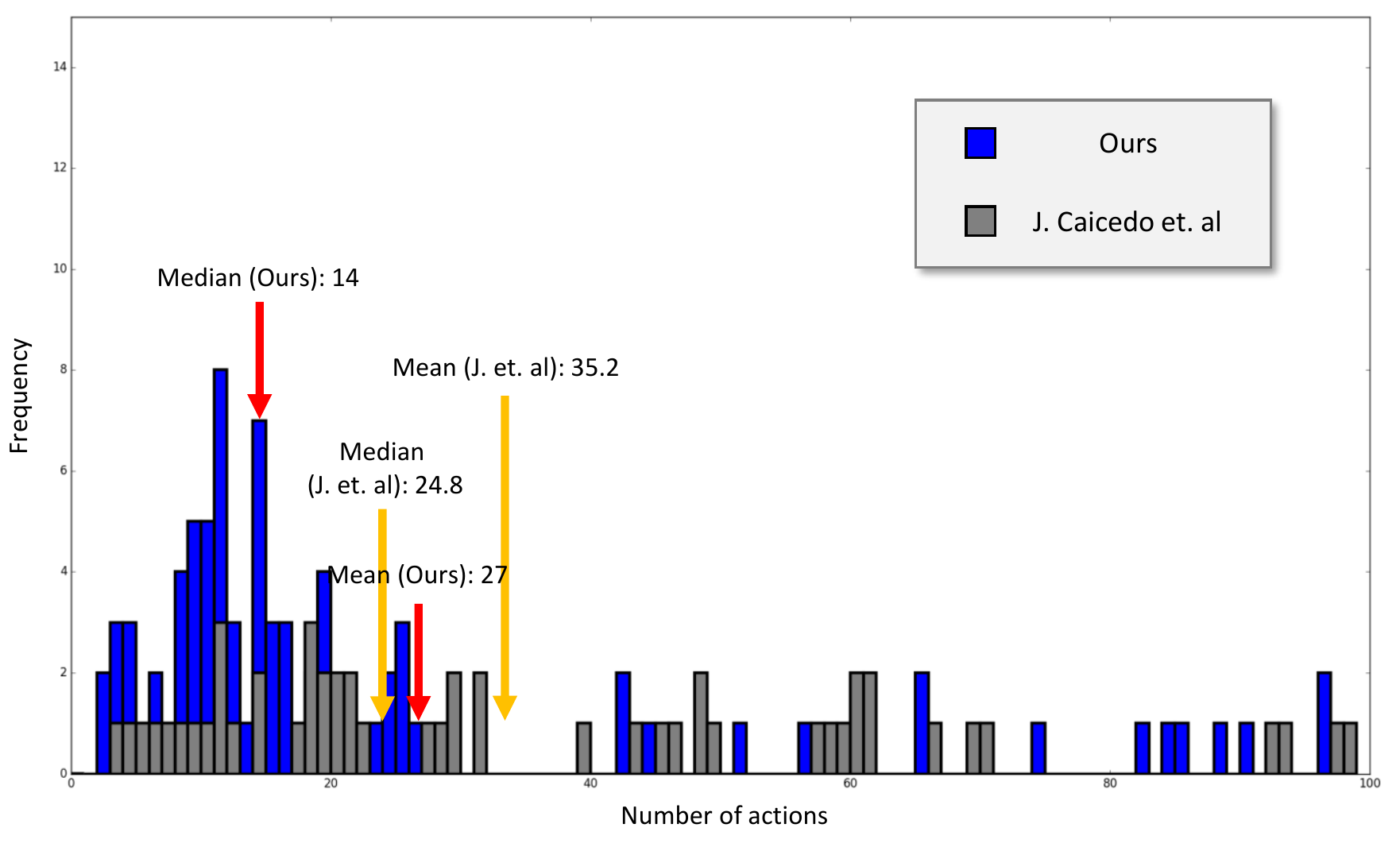}
		}
	}
	\caption{\small{Distribution of detections explicitly marked by the agent as a function of the number of actions required to reach the object. For each action, one region in the image needs to be processed. Both the single agent and our joint agent model can obtain most detections using around several tens of actions only. But the joint model obtains more detections with fewer actions for both categories.}}
	\label{fig:recall2}
\end{figure}

In \cite{caicedo2015active} the authors claim that the single agent localization algorithm can achieve higher recall values when compared with state-of-the-art object proposal methods with limited number of box proposals. We followed their setup and performed the same test to our joint model. Our experiments demonstrate that the proposed multi-agent method has a high recall value when using less proposals. Following the evaluation methods of Hosang et al. \cite{Hosang2015Pami}, we compare the recall of ours with those of the single agent baseline \cite{caicedo2015active} as well as one state-of-the-art object proposal method, Edgebox \cite{zitnick2014edge}. The results are from the combination of \textit{person+bicycle (VOC)} which provides stable configurations. Figure \ref{fig:recall1} and Figure \ref{fig:recall2} provide detailed comparisons.

\setlength{\textfloatsep}{20pt}
\begin{figure}[t]
    \centering
    {
    \includegraphics[width=1\columnwidth]{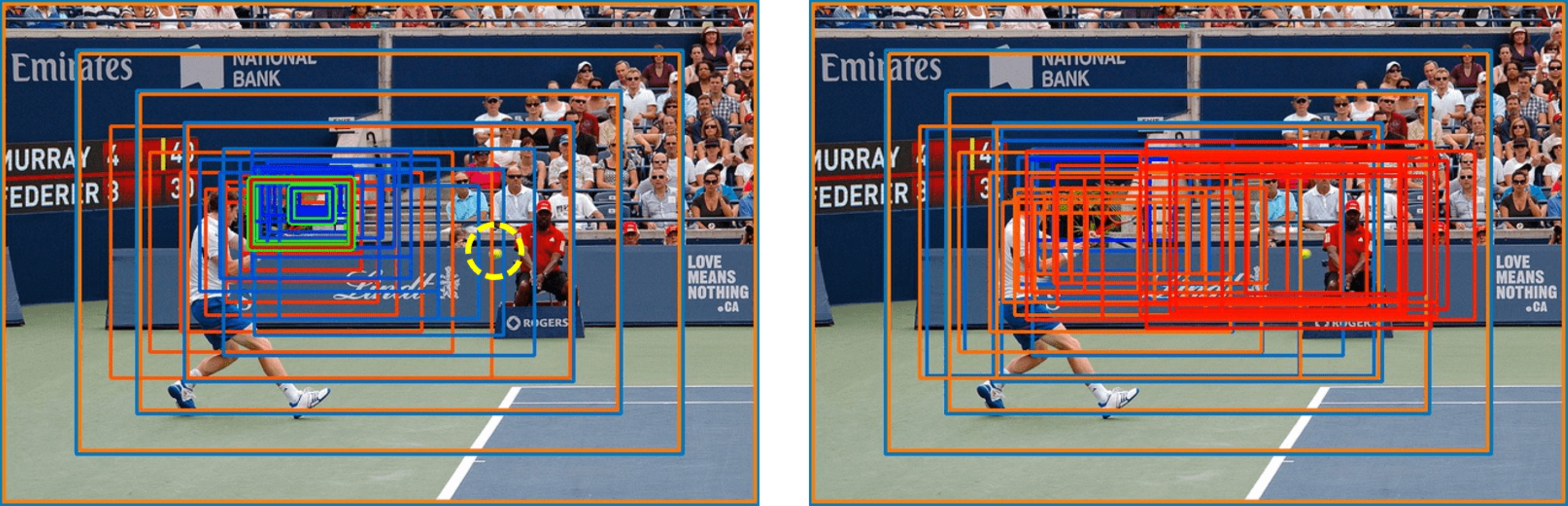}
    }
     \caption{\small{One failure case of joint detection. The true location of the tennis ball is highlighted with dashed yellow circle in the left image. Left: joint agent detection; right: single agent detection. }}
\label{fig:failure}
\end{figure}

\subsection{Failure Case Analysis}

In Figure \ref{fig:failure}, we show one interesting failure case of our method. In this case, our joint model correctly detects the racket but falsely locates a tennis ball on top of the racket. Meanwhile the true location of the ball is far away to the right. This phenomenon of over-fitting raises one important question. Does joint detection always help? 
The answer is clearly NO in general cases. Many combinations are not meaningful in the regard of joint detection. Actually, one can barely find shared images for totally unrelated object pairs such as, {\em e.g.} ``bird+car" etc. However, we did explore several more combinations that often coexist but have less spatial correlations. The results are demonstrated as follows.

\begin{table}[h] \small
	\caption{Localization accuracy. Top: single, bottom: joint.}
	\label{tab:test1}
	\begin{center}
		{
			\begin{tabular}{c|c|c|c|c|c|c|c}
				\hline
				\multicolumn{2}{c|}{(COCO)} & \multicolumn{2}{c|}{(COCO)} &
				\multicolumn{2}{c|}{(COCO)} & \multicolumn{2}{c}{(ImageNet)} \\
				\hhline{--------}
				\!\!fork\!\! & \!\!\!\!knife\!\!\!\! & \!\!\!\!oven\!\!\!\! & \!\!sink\!\!  &
				\!\!chair\!\! & \!\!\!\!tv\!\!\!\! & \!\!guitar\!\! &  \!\!mike\!\!  \\
				\hline
				31.9 &  45.2 & 38.2 & 34.3 & 35.1 & 57.1 & 80.9   & 45.4 \\
				\textbf{34.7} & \textbf{46.9}  & \textbf{42.4} & \textbf{37.7}  & 35.9 &  56.2  & \textbf{87.7} &  \textbf{50.2} \\
				\hline
			\end{tabular}
		}
	\end{center}
\end{table}

We noticed that even though such pairs do not display a fixed spatial correlation, they often have several major configurations of coexisting patterns. Therefore we can still consistently achieve better performance than single agent models, showcasing that meaningful messages were learned. The pair of ``chair+tv" is the least of this case and the positions of chairs and televisions in the images seem rather random even though they often coexist. In this setting, our joint model achieved similar performance with single models. This phenomenon shows that when no clear collaborative information exists, our proposal can perform as well as single agent models without messing up. We attribute this property to the gating mechanism by design.

\section{Conclusion}
\label{sec:concl}


Joint search of multiple objects under interaction often provides contextual cues to each other. By treating each detector as an agent, we present the first collaborative multi-agent deep reinforcement learning method that effectively learns the optimal policy for joint active object localization. Our technical contributions lie in the learnable cross Q-network communications and the joint exploitation sampling strategy. More importantly, we make a first stab to validate the concept of collaborative object localization by devising a computational model, which reveals interesting and intuitive co-detection patterns.

\newpage

\small{
\bibliographystyle{plain}
\bibliography{refr}
}

\end{document}